\documentclass{article}
\usepackage{ijcai21}

\usepackage{times}
\usepackage{soul}
\usepackage{url}
\usepackage[hidelinks]{hyperref}
\usepackage[utf8]{inputenc}
\usepackage[small]{caption}
\usepackage{graphicx}
\usepackage{amsmath}
\usepackage{amsthm}
\usepackage{booktabs}
\usepackage{algorithm}
\usepackage{algorithmic}
\usepackage{multirow}
\usepackage{amsmath,amssymb}
\usepackage{bbding}
\usepackage{dsfont}
\usepackage{amssymb}
\usepackage{pifont}
\newcommand{\cmark}{\ding{51}}%
\newcommand{\xmark}{\ding{55}}%
\title{A Sequence-to-Set Network for Nested Named Entity Recognition}

\author{
Zeqi Tan\footnote{Equal contribution.}
\and
Yongliang Shen$^\ast$
\and
Shuai Zhang
\and
Weiming Lu\footnote{Corresponding author.}
\and
Yueting Zhuang

\affiliations
College of Computer Science and Technology, Zhejiang University\\
\emails
\{zqtan, syl, zsss, luwm, yzhuang\}@zju.edu.cn
}

\begin{document}

\maketitle

\begin{abstract}
Named entity recognition (NER) is a widely studied task in natural language processing. Recently, a growing number of studies have focused on the nested NER. The span-based methods, considering the entity recognition as a span classification task, can deal with nested entities naturally. But they suffer from the huge search space and the lack of interactions between entities. To address these issues, we propose a novel sequence-to-set neural network for nested NER. Instead of specifying candidate spans in advance, we provide a fixed set of learnable vectors to learn the patterns of the valuable spans. We utilize a non-autoregressive decoder to predict the final set of entities in one pass, in which we are able to capture dependencies between entities. Compared with the sequence-to-sequence method, our model is more suitable for such unordered recognition task as it is insensitive to the label order. In addition, we utilize the loss function based on bipartite matching to compute the overall training loss. Experimental results show that our proposed model achieves state-of-the-art on three nested NER corpora: ACE 2004, ACE 2005 and KBP 2017.
The code is available at \url{https://github.com/zqtan1024/sequence-to-set}.
\end{abstract}

\section{Introduction}
\begin{figure}[h]
  \centering
  \includegraphics[width=\linewidth]{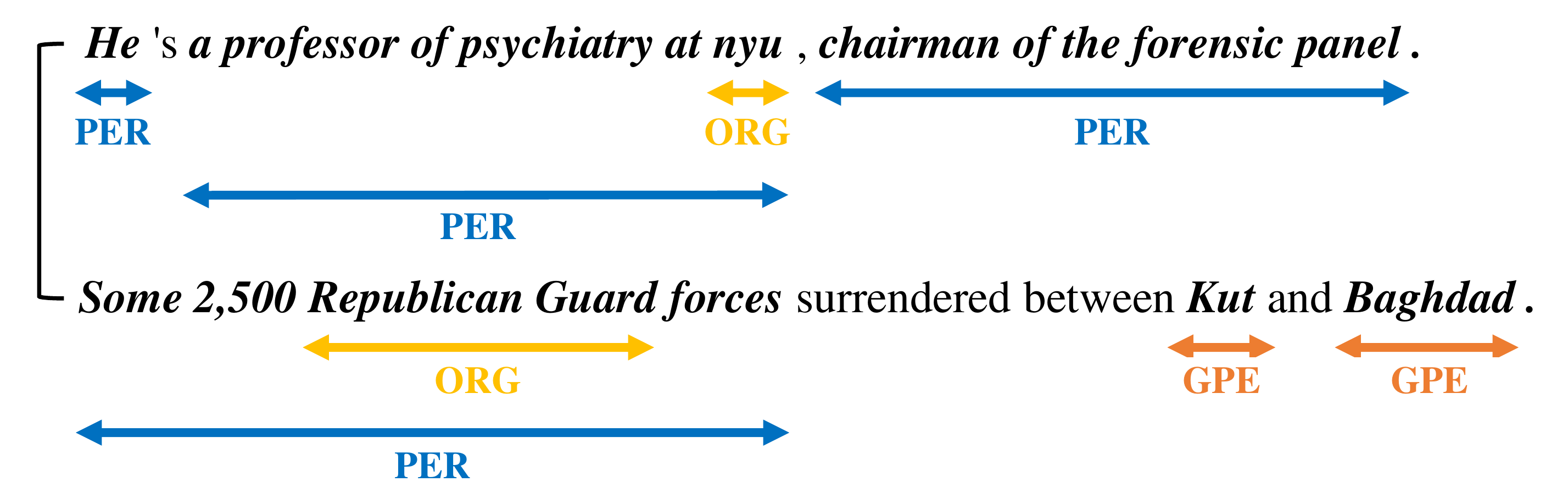}
  \caption{Examples for \textit{nested} entities from ACE05 corpora.}
   \label{fig:example}
\end{figure}

Named entity recognition (NER) is a fundamental task in natural language processing because named entities often contain the key information \cite{lample2016neural}. Nested named entities refer to that internal named entities are contained in external named entities as in Figure \ref{fig:example},  in which ``\textit{nyu}" and ``\textit{a professor of psychiatry at nyu}" are both named entities. Such nested entities are common in practice. For example, about 38\% of the entities in the ACE 2005 training dataset are nested. 
Traditional work \cite{huang2015bidirectional,lample2016neural,chiu2016named} models named entity recognition as a sequence labeling task, thus enabling to leverage RNN-based methods to recognize flat entities. However, since tokens in nested entities may belong to multiple labels, the traditional sequence labeling approaches cannot meet the demand.

Various methods have been proposed to deal with the nested NER task, including
the sequence-to-sequence methods \cite{ju2018neural,strakova2019neural,jue2020pyramid} and the span-based methods \cite{sohrab2018deep,zheng2019boundary,yu2020named}. The sequence-to-sequence method treats nested NER as a sequence generation task in which labels are decoded one by one in order. However, in the named entity recognition task, the output labels are essentially an unordered set. As shown in Figure \ref{fig:example}, there is apparently no ordinal relationship between the two GPE entities (``\textit{Kut}" and ``\textit{Baghdad}"). \citep{strakova2019neural} generates the output labels sequentially which is sensitive to the label order. In this manner, even if the model predicts all correct labels, it may cause an unreasonable training loss as a result of inconsistent order.
The span-based models classify the candidate spans which are extracted from a text sequence through various approaches. 
\citep{sohrab2018deep} enumerates all possible spans within a limit length and then predicts their classes. Unlike the exhaustive method, \citep{zheng2019boundary} first identifies the left and right boundaries separately and then matches them to form candidate spans. Similarly, ARN \cite{lin2019sequence} recognizes the spans of interest based on previously identified anchor words. Recently, \citep{yu2020named} scores candidate spans via the biaffine model \cite{DBLP:journals/corr/DozatM16} and  reaches the state-of-the-art. Although the above span-based methods achieve good performances, they also face some difficulties. The span-selecting methods face the problem of error propagation, because the boundaries or anchors are tend to be identified incorrectly, while the span-enumerating methods need to search for all possible regions. Besides, the candidate spans in these methods do not interact directly with each other and thus inappropriately ignore the dependencies between named entities.

To address the above issues, we propose a novel model that treats the named entity recognition as a sequence-to-set task. Considering that the entities in a sequence are inherently unordered, we expect to predict the entire set of entities in one pass. Therefore our model is no longer as sensitive to the label order as the sequence-to-sequence model. Compared to the span-based methods, instead of specifying candidate spans in advance, we provide a fixed set of learnable vectors that go through the entire train data to learn the patterns of the valuable spans. Hence we do not need to search for all possible spans anymore. Specifically, our model consists of three components: a sequence encoder, an entity set decoder and a loss function based on bipartite matching. We utilize BERT \cite{devlin-etal-2019-bert} and BiLSTM \cite{huang2015bidirectional} to construct the sequence encoder. In order to predict all entities of the sequence in one pass, we design a non-autoregressive entity set decoder. It receives the sequence encoding and a set of learnable vectors, which are called entity queries, to decode the final entity set. Furthermore, our decoder is able to capture the dependencies between entities quite naturally through a self-attention mechanism between entity queries. At the end, we utilize the loss function based on bipartite matching to compute the overall training loss.

Our main contributions are as follow:

\begin{itemize}
    \item We propose a novel sequence-to-set network for the prediction of the entity set. To the best of our knowledge, we are the first to consider named entity recognition as a sequence-to-set task. In addition, we predict the final entity set in one pass, while the sequence-to-sequence model predicts entities one by one. Since entities are inherently unordered, our model which is insensitive to the label order achieves a better performance.
    \item We provide a fixed set of entity queries to replace the explicit candidate spans, thus eliminating the need to search for all possible spans. Besides, 
    we are able to capture the dependencies between entities by using a self-attention mechanism which performs direct interactions between entity queries.
    \item Experimental results show that our model achieves state-of-the-art 
    on three widely used datasets, and outperforms several competing baseline models on F1 score by 0.56\% on ACE 2004, 1.65\% on ACE 2005
    and 2.99\% on KBP 2017.
\end{itemize}

\section{Model}
\begin{figure*}[h]
  \centering
  \includegraphics[width=\linewidth]{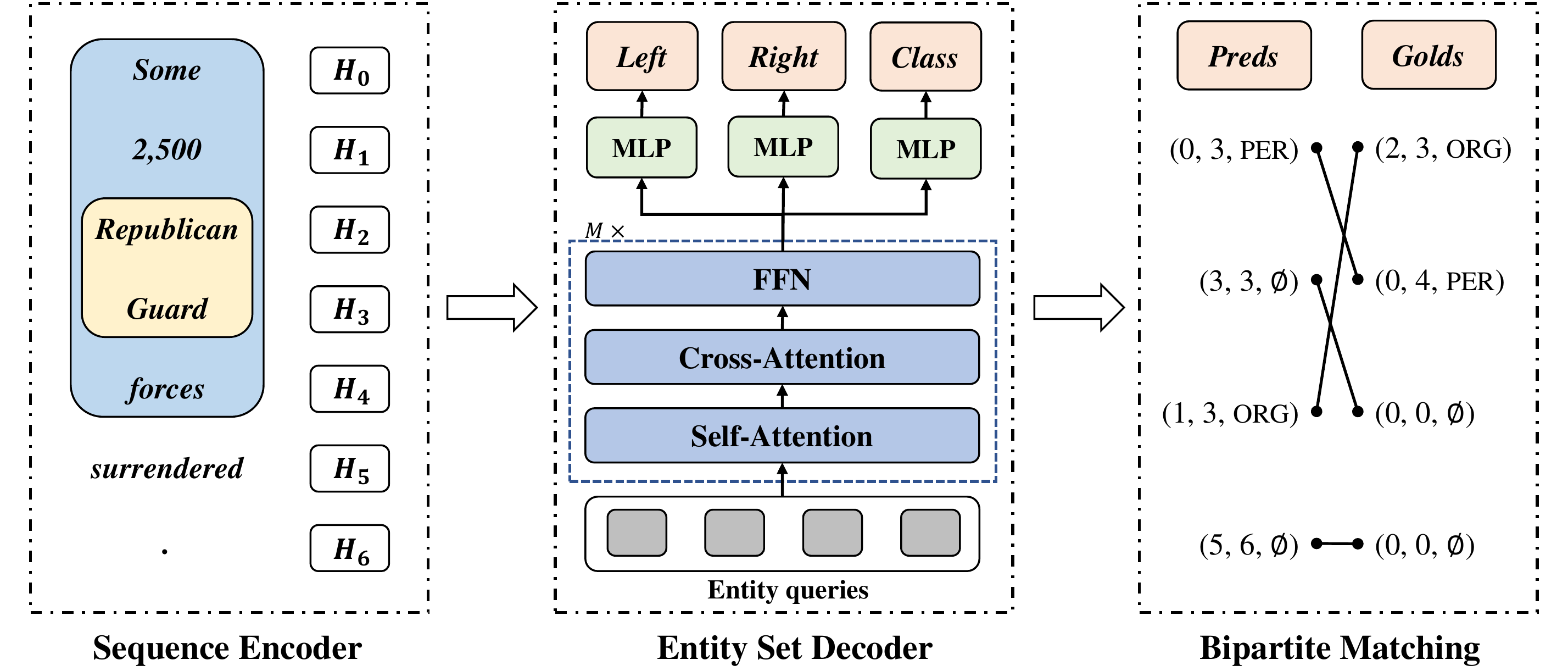}
  \caption{The architecture of our Sequence-to-Set Network. The representation of each token in sentence ``\textit{Some 2,500 Republican Guard forces surrendered.}" is fed into the entity set decoder along with the entity queries.
  Then the decoder transforms the queries to predicted entities. Finally, we score them by the loss function based on bipartite matching.}
   \label{fig:overview}
\end{figure*}

In order to transform the input sequence to the entity set, our model has three necessary components: a sequence encoder, an entity set decoder and a loss function based on bipartite matching. Firstly, the sequence encoder obtains rich contextual information from the input sentence. The entity set decoder then receives the sequence representation from the encoder and a set of entity queries to decode the left and right boundaries and categories of the predicted entity set. The bipartite matching is utilized to assign a unique prediction to each target entity so that the prediction loss of the entire model can be computed. The overview of our sequence-to-set model is shown in Figure \ref{fig:overview}. In the following subsections, we will introduce them in more detail.

\subsection{Sequence Encoder}
We use BERT \cite{devlin-etal-2019-bert} and BiLSTM \cite{huang2015bidirectional} to construct our sequence encoder. Given a input sequence,  we initially represent its $i$-th token by concatenating the BERT contextualized embeddings $x^{bert}_i$, the GloVE \cite{pennington-etal-2014-glove} embeddings $x^{glove}_i$, part-of-speech (POS) embeddings $x^{pos}_i$ and character-level embeddings $x^{char}_i$ together. We follow \citep{yu2020named} to obtain the contextualized embeddings $x^{bert}_i$ by encoding a target token with one surrounding tokens each side. The character-level embeddings are produced by a BiLSTM module which is the same as \citep{lample2016neural}. Then, the token representations $\left\{x_i\right\}$ are feed into another BiLSTM to get the final sequence representation $\operatorname{H} \in \mathbb{R}^{l \times d}$ as:

\begin{equation}
x_{i}=x_{i}^{bert} \oplus x_{i}^{glove} \oplus x_{i}^{pos} \oplus x_{i}^{char},
\end{equation}
\begin{equation}
\overrightarrow{\operatorname{H}}_{i}=\operatorname{LSTM}_{f}\left(x_{i}, \overrightarrow{\operatorname{H}}_{i-1}; \theta_f\right),
\end{equation}
\begin{equation}
\overleftarrow{\operatorname{H}}_{i}=\operatorname{LSTM}_{b}\left(x_{i}, \overleftarrow{\operatorname{H}}_{i+1}; \theta_b\right),
\end{equation}
\begin{equation}
\operatorname{H}_{i}=\overrightarrow{\operatorname{H}}_{i} \oplus \overleftarrow{\operatorname{H}}_{i},
\end{equation}
where $l$ is the sequence length and $d$ is twice the hidden size of LSTM, $\oplus$ denotes the concatenation operation and $\theta_f$ and $\theta_b$ denote the parameters of the forward and backward LSTM. The $\overrightarrow{\operatorname{H}}_{i}$ and $\overleftarrow{\operatorname{H}}_{i}$ are the hidden states at the position $i$ of the forward and backward LSTM.

\subsection{Entity Set Decoder}
The entity set decoder follows the classical transformer framework and uses self-attention as well as cross-attention mechanisms to transform $N$ entity queries. Different from the sequence-to-sequence method \cite{strakova2019neural} to predict entities one by one, we aim to decode all entities of a sequence in parallel due to the inherent disorder of entities. Similar to \cite{gu2018non}, we use a non-autoregressive decoder thereby allowing us to obtain $N$ predictions in one pass. Through the self-attention mechanism between entity queries, the decoder is able to capture the dependencies between entities. The decoder can also effectively acquire contextual information via the cross-attention mechanism. Because the decoding manner is non-autoregressive, we do not need to adopt the masking mechanism to prevent information leakage and thus can obtain the complete contextual semantic information. In addition, since the set of entity queries is used as the input for each sequence during the training process, we have an access to the global view of the entire dataset.

The $N$ entity queries denoted by $\mathbf{Q}_{span}$ are transformed into output embeddings by the entity set decoder of $M$ layers. This process mainly involves the multi-head attention mechanism \cite{vaswani2017attention}. For simplicity, we denote the attention as the following equation:

\begin{equation}
\text { Attention }(\mathbf{Q}, \mathbf{K}, \mathbf{V})=\operatorname{softmax}\left(\frac{\mathbf{Q K}^{T}}{\sqrt{d_{k}}}\right) \mathbf{V},
\end{equation}
where $\mathbf{Q}$, $\mathbf{K}$, $\mathbf{V}$ are the query matrix, key matrix and value matrix respectively, and the $1 / \sqrt{d_{k}}$ is the scaling factor. In the self-attention, $\mathbf{Q}=\mathbf{K}=\mathbf{V}=\mathbf{Q}_{span}$. And in the cross-attention, $\mathbf{Q} = \mathbf{Q}_{span}$ while $\mathbf{K} = \mathbf{V} = \mathbf{H}$. The multi-head attention can be formulated as follows:

\begin{equation}
\text { head }_{i}=\text {Attention}\left(\mathbf{Q} \mathbf{W}_{i}^{Q}, \mathbf{K} \mathbf{W}_{i}^{K}, \mathbf{V} \mathbf{W}_{i}^{V}\right),
\end{equation}

\begin{equation}
\mathbf{R}=\operatorname{Concat}\left(\operatorname{head}_{1}, \ldots, \text { head }_{h}\right) \mathbf{W}^{O},
\end{equation}
where $\mathbf{W}_{i}^{Q}$, $\mathbf{W}_{i}^{K}$, $\mathbf{W}_{i}^{V}$ and $\mathbf{W}^{O}$ are trainable projection parameters, and $h$ is the number of head. We take $\mathbf{R}$ computed by the cross-attention module as the input of FFN.

The FFN module is composed of a 3-layer perceptron with ReLU activation function and a linear projection layer. Through FFN, we denote the final output embeddings as $\operatorname{U} \in \mathbb{R}^{N \times d}$. We add an additional label $\varnothing$ to indicate that no entity is recognized because we predict a fixed-size set of $N$ entities, where $N$ is set larger than the actual number of entities in a sequence. Given an entity query $u \in \mathbb{R}^{d}$ in $\operatorname{U}$, the classification process can be defined by:

\begin{equation}
    p^c = \operatorname{MLP}_c(u),
\end{equation}

\begin{equation}
    \operatorname{H}_{fuse} = \operatorname{dup}(u,l) \oplus \operatorname{H},
\end{equation}

\begin{equation}
    p^l = \operatorname{MLP}_l(\operatorname{H}_{fuse}),
\end{equation}

\begin{equation}
    p^r = \operatorname{MLP}_r(\operatorname{H}_{fuse}),
\end{equation}
where $p^c$, $p^l$ and $p^r$ denote the probability of classification for classes, left boundaries and right boundaries respectively. $\oplus$ denotes the concatenation operation and $\operatorname{MLP}$ denotes the multilayer perceptron with softmax function in the last layer. The function $\operatorname{dup}$ will duplicate $u$ for $l$ times into shape $\mathbb{R}^{l \times d}$.

\subsection{Bipartite Matching}
After decoding the $N$ predictions, the main difficulty in training is to score the predicted entities (\textit{left}, \textit{right}, \textit{class}) according to the golden entities. To cope with the problem, we design a loss function based on bipartite matching. Before calculating the training loss, we first need to find a optimal matching between the predicted entity set and the golden entity set. We denote the golden set of entities by $y$, and the set of $N$ predictions by $\hat{y}=\left\{\hat{y}_{i}\right\}_{i=1}^{N}$. We also pad $y$ to the size of $N$ with $\varnothing$. To find the optimal matching we search for a permutation of $N$ elements $\beta \in \mathcal{O}_{N}$ with the lowest cost: 

\begin{equation}
\hat{\beta}=\underset{\beta \in \mathcal{O}_{N}}{\arg \min } \sum_{i}^{N} \mathcal{L}_{\operatorname{match}}\left(y_{i}, \hat{y}_{\beta(i)}\right),
\end{equation}
where $\mathcal{L}_{\operatorname{match}}\left(y_{i}, \hat{y}_{\beta(i)}\right)$ is a pair matching cost between the golden entity $y_i$ and a prediction with index $\beta(i)$. We compute this optimal assignment efficiently by using the Hungarian algorithm \cite{kuhn1955hungarian}. Considering the left boundaries, the right boundaries and the classes of entities, each element $i$ of the golden entity set can be seen as a $y_{i}=\left(l_{i}, r_{i}, c_{i}\right)$. We define $\mathcal{L}_{\operatorname{match}}\left(y_{i}, \hat{y}_{\beta(i)}\right)$ as:

\begin{table*}
\centering
\begin{tabular}{lrrrrrrrrrrr}
\toprule
\multirow{2}{*}{Dataset Statistics}   & \multicolumn{3}{c}{ACE 2004}& \multicolumn{3}{c}{ACE 2005} & \multicolumn{3}{c}{KBP 2017} & \multicolumn{2}{c}{GENIA}  \\
 \cmidrule(lr){2-4}  \cmidrule(lr){5-7} \cmidrule(lr){8-10} \cmidrule(lr){11-12}  
& Train  & Dev & Test & Train  & Dev & Test & Train  & Dev & Test & Train   & Test  \\
\midrule
\# sentences &  6200 &  745 &  812 &  7194 &  969 &  1047 &  10546 &  545 & 4267 &  16692 &   1854 \\
\# with nested entities  &  2712 &  294 &  388 &  2691 &  338 &  320 &  2809 &  182 &  1223 &  3522 &   446 \\
avg sentence length &  22.50 &  23.02 &  23.05 &  19.21 &  18.93 &  17.2 &  19.62 &  20.61 &  19.26 &  25.35 &    25.99 \\
\# total entities &  22204 &  2514 &  3035 &  24441 &  3200 &  2993 &  31236 &  1879 &  12601 &  50509 &    5506 \\
\# nested entities &  10149 &  1092 & 1417  & 9389 &  1112 &  1118 &  8773 &  605 &  3707 &  9064 &    1199 \\
 nested percentage (\%) &  45.71 & 46.69 &  45.61 & 38.41 & 34.75 &  37.35 &  28.09 & 32.20 &  29.42 &  17.95 &    21.78 \\
\bottomrule

\end{tabular}
\caption{Statistics of the datasets used in the experiments.}
\label{tab:statistics}
\end{table*}

\begin{equation}
\begin{aligned}
\mathcal{L}_{\operatorname{match}}\left(y_{i}, \hat{y}_{\beta(i)}\right) &=-\mathds{1}_{\left\{c_{i} \neq \varnothing\right\}}\left[{p}_{\beta(i)}^{c}\left(c_{i}\right)\right.\\
&+{p}_{\beta(i)}^{l}\left(l_{i}\right) \\
&\left.+{p}_{\beta(i)}^{r}\left(r_{i}\right)\right].
\end{aligned}
\end{equation}
After we get the optimal matching $\hat{\beta}(i)$, we define the final loss $\mathcal{L}(y, \hat{y})$ as:

\begin{equation}
\begin{aligned}
\mathcal{L}(y, \hat{y})&=\sum_{i=1}^{N}\left\{-\log {p}_{\hat{\beta}(i)}^{c}\left(c_{i}\right)\right.\\
&+\mathds{1}_{\left\{c_{i} \neq \varnothing\right\}}\left[-\log {p}_{\hat{\beta}(i)}^{l}\left(l_{i}\right)\right.\\
&\left.\left.-\log {p}_{\hat{\beta}(i)}^{r}\left(r_{i}\right)\right]\right\}.
\end{aligned}
\end{equation}

\section{Experiments}

\begin{table}
\centering
\begin{tabular}{lccc}
\toprule
\multirow{2}{*}{Model}   & \multicolumn{3}{c}{ACE 2004}  \\
 \cmidrule(lr){2-4} 
& P  & R & F1  \\
\midrule

\citep{katiyar2018nested} 
&  73.60 &  71.80 &  72.70 \\
\citep{shibuya2019nested}   & 83.73 & 81.91 & 82.81\\
\citep{strakova2019neural}  & - & - &  84.40  \\
\citep{jue2020pyramid}       & 86.08  & 86.48  & 86.28     \\
\citep{yu2020named}       & 87.30  & 86.00  & 86.70     \\
\midrule
Ours    & 88.46  & 86.10  & \textbf{87.26}  \\

\bottomrule
\toprule
\multirow{2}{*}{Model}   & \multicolumn{3}{c}{ACE 2005}  \\
 \cmidrule(lr){2-4} 
& P  & R & F1  \\
\midrule
\citep{katiyar2018nested}  & 70.60 & 70.40 & 70.50 \\
\citep{lin2019sequence}       & 76.20  & 73.60 & 74.90  \\
\citep{luo-zhao-2020-bipartite}       & 75.00  & 75.20  & 75.10    \\
\citep{shibuya2019nested}   & 82.98  & 82.42  & 82.70 \\
\citep{strakova2019neural}  & - & - &  84.33  \\
\citep{jue2020pyramid}       & 83.95  & 85.39  & 84.66     \\
\citep{yu2020named}       & 85.20  & 85.60  & 85.40     \\
\midrule
Ours    & 87.48   & 86.63 & \textbf{87.05}  \\
\bottomrule
\toprule
\multirow{2}{*}{Model}   & \multicolumn{3}{c}{KBP 2017}  \\
 \cmidrule(lr){2-4} 
& P  & R & F1  \\
\midrule
\citep{DBLP:conf/tac/JiPZNMMC17}       & 76.20 & 73.00 & 72.80  \\
\citep{lin2019sequence}       & 77.70  & 71.80 & 74.60  \\
\citep{luo-zhao-2020-bipartite}       & 77.10 & 74.30 & 75.60  \\
\citep{li-etal-2020-unified}       & 80.97  & 81.12  & 80.97  \\
\midrule
Ours    & 84.91  & 83.04  & \textbf{83.96}  \\
\bottomrule
\toprule
\multirow{2}{*}{Model}   & \multicolumn{3}{c}{GENIA}  \\
 \cmidrule(lr){2-4} 
& P  & R & F1  \\
\midrule
\citep{lin2019sequence}       & 75.80  & 73.90 & 74.80  \\
\citep{luo-zhao-2020-bipartite}       & 77.40  & 74.60  & 76.00    \\
\citep{wang-etal-2020-hit}       & 78.10  & 74.40  & 76.20  \\
\citep{shibuya2019nested}   & 78.07 & 76.45 & 77.25 \\
\citep{strakova2019neural}  & - & - &  78.31 \\
\citep{jue2020pyramid}       & 79.45  & 78.94  & 79.19     \\
\citep{yu2020named}       & 81.80  & 79.30  & \textbf{80.50}     \\
\midrule
Ours    & 82.31 & 78.66  & 80.44  \\
\bottomrule

\end{tabular}
\caption{The overall performances for \textit{\textbf{nested}} NER tasks.}
\label{tab:nested}
\end{table}

\subsection{Experimental Setup}
In the experimental setup, we utilize the widely used ACE 2004, ACE 2005, KBP 2017 and GENIA datasets. For ACE 2004 and ACE 2005, we follow the previous work \cite{katiyar2018nested,lin2019sequence} to keep files from $\operatorname{bn}$, $\operatorname{nw}$ and $\operatorname{wl}$ and divide these files into train, dev and test sets in the ratio of 8:1:1, respectively. For GENIA, we use geniacorpus3.02 as in \citep{katiyar2018nested}. For KBP 2017, we use the 2017 English evaluation dataset and the same split strategy in \citep{lin2019sequence}. Table \ref{tab:statistics} shows the statistical results in detail for all the above datasets.

Standard precision, recall and F1-measure are employed as evaluation metrics. An entity is considered correct only if both the entity boundary and the entity label are correct. The checkpoint that has the best F1 score in the development set is chosen to evaluate the test set.

For comparison with \citep{jue2020pyramid} and \citep{yu2020named}, we use the cased large version of BERT in most experiments. We also adopt the GloVe \cite{pennington-etal-2014-glove} 100-dimension pre-trained word vectors. The character-level BiLSTM layer is set to 1 and the token-level BiLSTM layer is set to 3. The hidden size of the former is 50, and the hidden size of the latter is half of the BERT hidden size. Specifically, on the GENIA dataset, we replace BERT with BioBERT \cite{10.1093/bioinformatics/btz682}, GloVE with BioWordvec \cite{chiu-etal-2016-train}. The number of entity queries $N$ is set to 60 and the vectors are randomly initialized with the normal distribution $\mathcal{N}(0.0, 0.02)$. The number of our decoder layer $M$ is set to 3. The number of the attention heads is set to 8. The number of the MLP layer is set to 1. We use the AdamW \cite{DBLP:journals/corr/abs-1711-05101} optimizer with a linear warmup-decay learning rate schedule (with peak learning rate of 2e-5 and epoch of 100), a dropout with the rate of 0.1 and a batch size of 8.

\subsection{Baselines}
We compare our model with several start-of-the-art methods on ACE 2004, ACE 2005, KBP 2017 and GENIA datasets:
\begin{itemize}
    \item \textbf{Biaffine:} \citep{yu2020named} utilizes the biaffine model to score pairs of start and end tokens in a sequence.
    \item \textbf{Pyramid:} \citep{jue2020pyramid} designs the normal and inverse pyramidal structures to identify entities through bidirectional interactions.
    \item \textbf{BiFlaG:} \citep{luo-zhao-2020-bipartite} proposes a bipartite flat-graph network with two interacting subgraph modules.
    \item \textbf{HIT:} \citep{wang-etal-2020-hit} designs head-tail detector and token-interaction tagger to express the nested structure.
    \item \textbf{Seq2seq:} \citep{strakova2019neural} considers the nested NER as a sequence-to-sequence problem. The encoded labels are predicted one by one by the decoder.
    \item \textbf{Second-best:} \citep{shibuya2019nested} searches a span of each extracted entity for nested entities with second-best sequence decoding.
    \item \textbf{ARN:} \citep{lin2019sequence} leverages the head-driven phrase structures to predict nested named entities.
    \item \textbf{Hyper-Graph:} \citep{katiyar2018nested} makes use of the BILOU tagging scheme to learn the hypergraph representation.
    \item \textbf{KBP17-Best:} \citep{DBLP:conf/tac/JiPZNMMC17} reports the best previous results for nested NER tasks.
\end{itemize}

We don't compare our model with BERT-MRC \cite{li-etal-2020-unified}, because it uses additional external resources to construct the questions, which essentially introduces descriptive information about the categories. 

\subsection{Overall Performances}
The overall performances of the proposed model against baselines for nested NER tasks on ACE 2004, ACE 2005, KBP 2017 and GENIA datasets are shown in Table \ref{tab:nested}. 
Our proposed model outperforms the state-of-the-art model on ACE 2004, ACE 2005 and KBP 2017 datasets.
To be specific, the F1 scores of our model improve by +0.56\%, +1.65\% and +2.99\% over previous SOTA performances on ACE 2004, ACE 2005 and KBP 2017, respectively. Meanwhile, we achieve the comparable result with the SOTA model on GENIA dataset.
The experimental results demonstrate the ability of our model to recognize named entities in the sequence. 

The main improvement of the proposed model comes from the fact that we treat the nested NER as a sequence-to-set problem, which is consistent with the inherent disorder of entities. Compared with the sequence-to-sequence model, our model is insensitive to the label order and achieves a better performance.

In addition, in order to choose an appropriate number of entity queries, we design a comparison experiment on ACE 2005 with a maximum number of entities of 27. The evaluation results are shown in Table \ref{tab:queries}. We observe that when the number is obviously larger than the ground truth, the performance of the model does decrease. Eventually, the query number $N$ in our experiments is set to 60.

\begin{table}
\centering
\begin{tabular}{cccc}
\toprule
\multirow{2}{*}{\# Entity Query}   & \multicolumn{3}{c}{ACE 2005}  \\
 \cmidrule(lr){2-4} 
& P  & R & F1  \\
\midrule
40       & 86.84 & 86.93 & 86.88  \\
60       & 87.48  & 86.63 & \textbf{87.05}  \\
80       & 87.13 & 86.23 & 86.68  \\
100      & 86.28  & 86.60  & 86.44  \\
200      & 86.63  & 85.53  & 86.08  \\
\bottomrule

\end{tabular}
\caption{The performances with different number of entity queries.}
\label{tab:queries}
\end{table}

\begin{table}
\centering
\begin{tabular}{ccccc}
\toprule
\multirow{2}{*}{\# Layer} & \multirow{2}{*}{Interaction}  & \multicolumn{3}{c}{ACE 2005}  \\
 \cmidrule(lr){3-5} 
& & P  & R & F1  \\
\midrule
1  & \xmark     & 82.14 & 82.85 & 82.49  \\
1  & \cmark     & 86.64  & 84.12 & 85.36  \\
2  & \xmark     & 83.57 & 83.46 & 83.51  \\
2  & \cmark     & 85.80 & 86.63 & 86.21  \\
3  & \xmark     & 85.26 & 85.23 & 85.24  \\
3  & \cmark & 87.48 & 86.63 & \textbf{87.05}  \\

\bottomrule

\end{tabular}
\caption{Ablation studies for decoder layer number and interaction between entity queries.}
\label{tab:ablation}
\end{table}

\subsection{Ablation Studies}
To demonstrate the effectiveness of the several modules in our method, we conduct a series of ablation studies on ACE 2005. First, according to our intuition, the more decoder layers we stack, the more expressive the model is. Table \ref{tab:ablation} verifies this point: when we reduce the number of decoder layer from 3 to 2 and from 2 to 1, the model performance decreases by 0.84\% and 0.85\%, respectively. We also conduct experiments with 6 layers, but the model performance does not improve obviously. Furthermore, from Table \ref{tab:ablation} we also observe that regardless of the number of layers, the performance of the model consistently decreases with the absence of the interaction (\textit{self attention}) between entity queries. When the number of decoder layer is 3, 2 and 1, the results decrease by 1.81\%, 2.70\% and 2.87\% respectively. This is because the decoder can capture the dependencies between entities by using a self-attention mechanism which performs direct interactions between entity queries.

Likewise, we explore the learning ability of entity queries and the effectiveness of bipartite matching loss. In Table \ref{tab:ablation_}, $\operatorname{Freeze \; Queries}$ means we keep the parameters of the entity queries unchanged, and $\operatorname{CE \; Loss}$ refers to replace bipartite matching loss with cross-entropy loss. We can see that freezing the entity queries results in the performance drops of 0.40\% and 0.81\% on ACE 2004 and ACE 2005, respectively. This indicates that the entity queries do learn the patterns of the valuable regions. Besides, we can gain improvement of 2.67\% and 1.11\% on ACE 2004 and ACE 2005 respectively by replacing cross-entropy loss with bipartite matching loss, which shows the power of bipartite matching loss on the unordered prediction task. Compared with the sequence-to-sequence model, we can not only predict all named entities in one pass, but are able to utilize the more suitable bipartite matching loss for the unordered entity recognition task.

\begin{table}
\centering
\begin{tabular}{lccc}
\toprule
\multirow{2}{*}{Settings} & \multicolumn{3}{c}{ACE 2004}  \\
\cmidrule(lr){2-4}
  & P  & R & F1  \\
\midrule
Freeze Queries     & 87.37  & 86.36 & 86.86  \\
CE Loss            & 85.55  & 83.66 & 84.59  \\
Full Model         & 88.46  & 86.10  & \textbf{87.26}  \\
\bottomrule
\toprule
\multirow{2}{*}{Settings} & \multicolumn{3}{c}{ACE 2005}  \\
\cmidrule(lr){2-4}
  & P  & R & F1  \\
\midrule
 Freeze Queries     & 85.78 & 86.70 & 86.24  \\
 CE Loss            & 85.24  & 86.66 & 85.94  \\
 Full Model         & 87.48 & 86.63 & \textbf{87.05}  \\

\bottomrule

\end{tabular}
\caption{Ablation studies for the learnable entity queries and bipartite matching loss.}
\label{tab:ablation_}
\end{table}

\section{Related Work}
Recent nested named entity recognition methods can be categorized as sequence-based \cite{ju2018neural,jue2020pyramid}, hypergraph-based \cite{lu2015joint,wang2018neural,katiyar2018nested}, and span-based \cite{xu2017local,sohrab2018deep,zheng2019boundary,yu2020named} approaches. \citep{ju2018neural} proposed a layered model to recognize entities from inside to outside sequentially. Pyramid \cite{jue2020pyramid} were designed with normal and inverse pyramidal structures to identify entities through bidirectional interactions. \citep{lu2015joint} were the first to propose a hypergraph-based approach to cope with the nested entity mention detection problem. \citep{wang2018neural} utilized segmental hypergraph to model arbitrary combinations of mentions. \citep{katiyar2018nested} made use of the BILOU tagging scheme to learn the hypergraph representation.  The span-based models classify the candidate spans which are extracted from a text sequence through various approaches. Exhaustive Model \cite{sohrab2018deep} enumerated all possible spans in a sentence within a limited length and afterwards to predict their categories. \citep{luan2019general} used a dynamic span graphs to select the most confident entity spans.  Unlike the above methods, BERT-MRC \cite{li-etal-2020-unified} formulated the named entity recognition problem as a machine reading comprehension task so that the flat and nested cases can be handled uniformly. Other works \cite{lin2019sequence,fisher2019merge,strakova2019neural}  also presented various approaches to solve the nested NER problem.

We consider the nested named entity recognition as a sequence-to-set task, thus avoiding the error propagation problem of the sequence-based methods \cite{ju2018neural}. Compared to the span-based methods, we provide a fixed set of entity queries to learn the patterns of the valuable spans. Thus we do not need to search for all possible regions anymore. Besides, we predict the final set of entities in one pass, while the sequence-to-sequence method \cite{strakova2019neural} predicts entities one by one. Since the entities are inherently unordered, our model which is insensitive to the label order achieves a better performance. Similarly, DETR \cite{carion2020end} treats the object detection as a set prediction task. This end-to-end approach effectively eliminates the need for many manually designed components in the two-stage methods such as Faster R-CNN \cite{ren2016faster}.

\section{Conclusion}
In this paper, we propose a novel sequence-to-set network for nested NER. Due to the fact that the entities in a sequence are inherently unordered, we design a non-autoregressive decoder to generate the set of entities in one pass. To score the predicted entities respect to the golden entities, we design a bipartite matching loss which is more suitable for the unordered entity recognition task. Compared with the sequence-to-sequence model, our model is insensitive to the label order and achieves a better performance. Besides, we can capture the dependencies between entities by using a self-attention mechanism which performs direct interactions between entity queries. Experiments show that our model achieves state-of-the-art on 
ACE 2004, ACE 2005 and KBP 2017 datasets. In the future, how to represent the dependencies between entities better will be considered.

\section*{Acknowledgments}

This work is supported by the National Key Research and Development Project of China (No. 2018AAA0101900), the Chinese Knowledge Center of Engineering Science and Technology (CKCEST) and MOE Engineering Research Center of Digital Library.

\bibliographystyle{named}
\bibliography{ijcai21}

\end{document}